\documentclass[journal]{IEEEtran}

\IEEEoverridecommandlockouts                              

\newtheorem{definition}{Definition}
\newtheorem{proposition}{Proposition}
\newtheorem{lemma}{Lemma}
\newtheorem{theorem}{Theorem}

\usepackage{array} 
\usepackage{makecell} 
\usepackage{algorithm}  
\usepackage{algpseudocode}
\usepackage{multirow}
\usepackage{epsfig} 
\usepackage{amsmath} 

\usepackage{amsfonts} 

\usepackage[flushleft]{threeparttable}
\usepackage{graphicx}
\usepackage{cite} 

\usepackage{tabularx} 

\usepackage{subcaption}
\usepackage{hyperref}

\usepackage{tikz}
\usetikzlibrary{arrows.meta, positioning, fit}

\usepackage{color}

\title{
Safe Autonomous Lane Changing: Planning with Dynamic Risk Fields and Time-Varying Convex Space Generation
}

\author{
    Yijun Lu$^{1}$,
    Zhihao Lin$^{2}$,
    and Zhen Tian$^{2}$%
    \thanks{$^{1}$Yijun Lu is with Waseda University, Tokyo, Japan (e-mail: yijun@ruri.waseda.jp).}%
    \thanks{$^{2}$Zhihao Lin and Zhen Tian are with the James Watt School of Engineering, University of Glasgow, Glasgow, G12 8QQ, U.K. (e-mail: 2800400L@student.gla.ac.uk; tian.zhen@glasgow.ac.uk).}%
}

\begin{document}

\maketitle
\thispagestyle{empty}
\pagestyle{empty}

\begin{abstract}
This paper presents a novel trajectory planning pipeline for complex driving scenarios like autonomous lane changing, by integrating risk-aware planning with guaranteed collision avoidance into a unified optimization framework. We first construct a dynamic risk fields (DRF) that captures both the static and dynamic collision risks from surrounding vehicles. Then, we develop a rigorous strategy for generating time-varying convex feasible spaces that ensure kinematic feasibility and safety requirements. The trajectory planning problem is formulated as a finite-horizon optimal control problem and solved using a constrained iterative Linear Quadratic Regulator (iLQR) algorithm that jointly optimizes trajectory smoothness, control effort, and risk exposure while maintaining strict feasibility. Extensive simulations demonstrate that our method outperforms traditional approaches in terms of safety and efficiency, achieving collision-free trajectories with shorter lane-changing distances (28.59 m) and times (2.84 s) while maintaining smooth and comfortable acceleration patterns. In dense roundabout environments the planner further demonstrates robust adaptability, producing larger safety margins, lower jerk, and superior curvature smoothness compared with APF, MPC, and RRT based baselines. These results confirm that the integrated DRF with convex feasible space and constrained iLQR solver provides a balanced solution for safe, efficient, and comfortable trajectory generation in dynamic and interactive traffic scenarios.
\end{abstract}

\section{Introduction}
Autonomous lane changing (ALC) constitutes a fundamental yet demanding capability for high-level autonomous driving systems, serving as a prerequisite for operational efficiency and traffic flow stability~\cite{cremades2025classically,alyamani2023driver}. Unlike lane-keeping tasks that primarily rely on distinct road features for lateral stability, ALC requires the autonomous vehicle (AV) to actively navigate out of its current safe corridor into a dynamic, often contested, target space~\cite{mei2025driving}. This maneuver involves highly coupled longitudinal and lateral vehicle dynamics, which must be executed within the strict geometric limits of the road infrastructure while interacting with heterogeneous traffic participants~\cite{10186066}. In mixed autonomy scenarios, where AVs share the road with human-driven vehicles (HDVs), the challenge is further amplified; the AV must not only ensure its own kinematic feasibility but also anticipate and negotiate with surrounding drivers who exhibit diverse and often aggressive driving behaviors~\cite{10078438}.

The complexity of generating high-fidelity ALC trajectories arises from the intricate interplay of three critical technical challenges. 
First, the decision-making environment is inherently stochastic and interactive. The motion of surrounding vehicles is not merely a physical state but a manifestation of latent driver intentions, requiring the AV to infer probabilistic future behaviors in real-time~\cite{peng2025diffusion, 10486836}. 
Second, the trajectory planning problem is mathematically governed by complex non-convex constraints. The requirement to avoid dynamic obstacles creates a disjoint, time-varying free space, making it computationally arduous to guarantee a globally optimal solution without getting trapped in local minima~\cite{9737041}. 
Third, the planning framework must reconcile conflicting multi-objective criteria. It must achieve a pareto-optimal trade-off between aggressive maneuvering for efficiency, smooth control for passenger comfort, and conservative margins for safety, a balance that is difficult to maintain under rapidly changing traffic conditions~\cite{10382432}.

To address these challenges, existing literature generally categorizes solutions into potential field, learning-based, and optimization-based methods, yet each faces distinct limitations. 
Potential field methods~\cite{vogel2022you, gao2024trajectory}, while computationally efficient, often suffer from local minima and fail to explicitly handle non-holonomic kinematic constraints~\cite{KEMPF2024111782}. 
Learning-based approaches, such as Reinforcement Learning (RL)~\cite{zhu2023bi, Chen2024}, excel in capturing complex interaction patterns but typically lack the interpretability and rigorous safety guarantees required for safety-critical systems~\cite{Ladosz2024}. 
Conversely, optimization-based methods (e.g., MPC)~\cite{9693175} provide constraint satisfaction but struggle with the high computational burden of solving non-convex optimization problems in highly dynamic environments~\cite{9658226}. 
Consequently, there remains a pressing need for a framework that unifies rigorous constraint satisfaction with risk-aware efficiency for real-time applications.

\begin{figure}[t]
    \centering
    \includegraphics[width=1\linewidth]{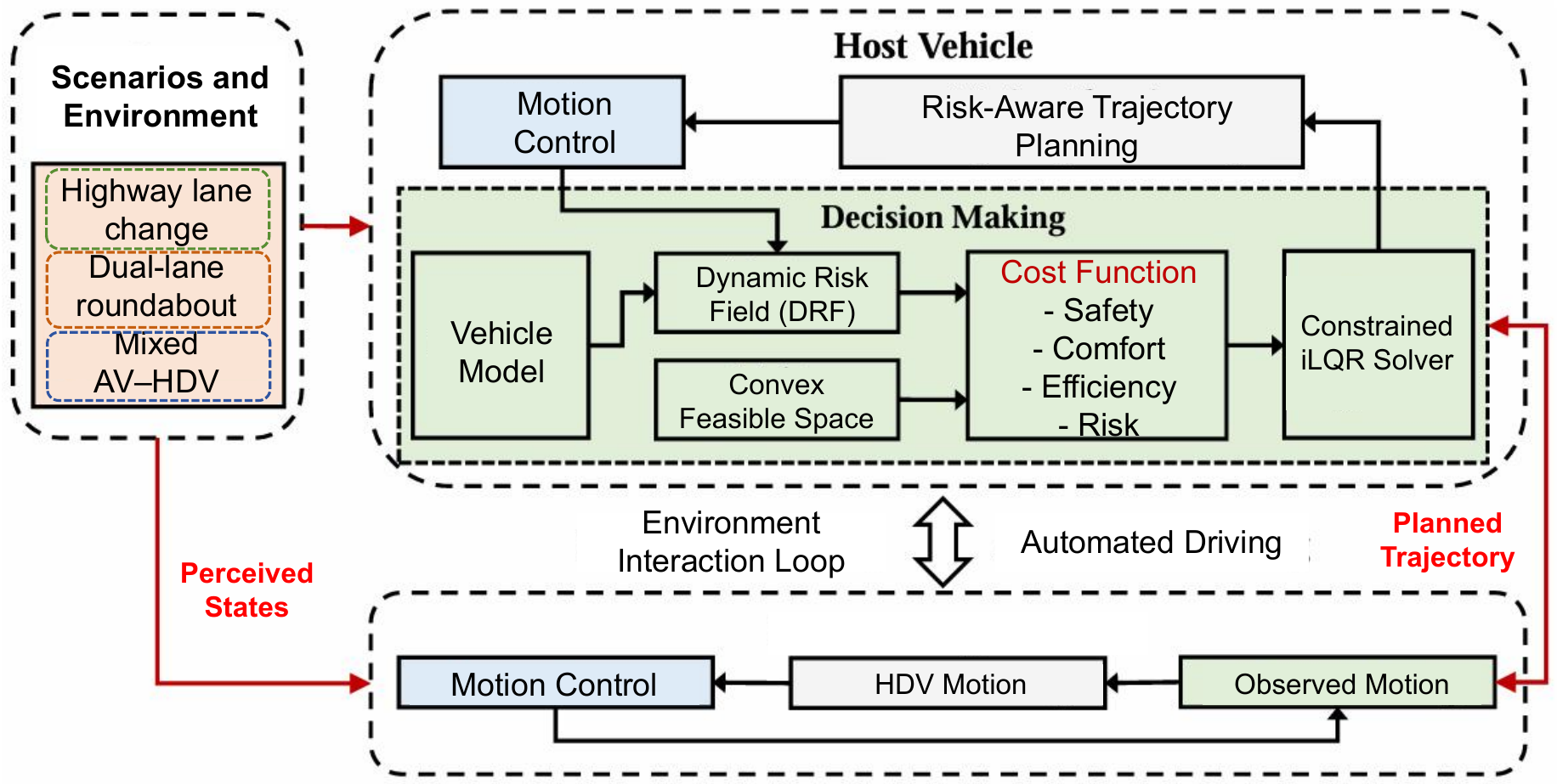}
    \vspace{-2mm}
    \caption{
    System architecture of the proposed risk-aware trajectory planning framework.
    }
    \label{fig:system_framework}
    \vspace{-2mm}
\end{figure}

In this work, we propose a unified trajectory planning framework that synergizes Dynamic Risk Fields (DRF) with time-varying convex feasible spaces. As illustrated in Fig.~\ref{fig:system_framework}, our approach decomposes the complex lane-changing problem into manageable sub-problems. 
Instead of treating obstacles as simple geometric boundaries, we construct a DRF that encodes collision likelihood using both static and dynamic risk terms. Simultaneously, to overcome the non-convexity of obstacle avoidance, we generate a Convex Feasible Space (CFS) that defines a safe, distinct corridor respecting the host vehicle’s kinematic limits. 
These components are integrated into a Constrained iterative Linear Quadratic Regulator (iLQR) solver, which generates optimal trajectories that are both risk-minimizing and dynamically feasible.
The main contributions of this paper are summarized as follows:

\begin{itemize}
    \item We propose a comprehensive Dynamic Risk Field (DRF) model that unifies static environmental constraints and dynamic collision risks, providing a continuous and differentiable metric for interaction-aware safety assessment.
    
    \item We develop a sequential convexification strategy that constructs time-varying convex feasible spaces. This guarantees kinematic feasibility and transforms the non-convex collision avoidance problem into a tractable form for real-time solving.
    
    \item We formulate the planning task as a finite-horizon optimal control problem solved via a Constrained iLQR algorithm. This approach jointly optimizes trajectory smoothness, control effort, and risk exposure while strictly enforcing safety constraints.
\end{itemize}

The remainder of this paper is organized as follows: Section II discusses related works. Section III details the methodology including DRF and convex feasible spaces. Section IV presents the optimization problem formulation and solution. Section V provides experimental analysis, followed by the conclusion in Section VI.

\section{Related Work}
Existing methodologies for autonomous lane changing are extensive and diverse. To contextualize our proposed framework, we categorize relevant literature into three distinct streams: (A) risk assessment and field-theoretic methods, (B) optimization-based planning with convexification, and (C) learning-based and hybrid control strategies.

\subsection{Risk Assessment and Field-Theoretic Methods}
Accurate quantification of collision risk is the cornerstone of safe trajectory planning. Traditional Artificial Potential Fields (APF) construct virtual force fields where obstacles repel the ego vehicle and target lanes attract it~\cite{vogel2022you}. While APF offers computational efficiency and intuitive obstacle avoidance, it inherently suffers from local minima issues and often yields oscillatory paths that disregard vehicle kinematic constraints~\cite{gao2024trajectory}.

To overcome these geometric limitations, recent research has evolved from static potential fields to dynamic "Risk Fields." These approaches integrate physical properties—such as velocity and mass—into the field generation. For instance, Wang et al.~\cite{wang2023risk} proposed a driving risk field that accounts for the kinetic energy of surrounding vehicles, effectively mapping the severity of potential collisions into a scalar cost. Similarly, Nie et al.~\cite{nie2023interaction} integrated interaction-aware potential fields with Model Predictive Control (MPC) to better balance risk avoidance with dynamic feasibility. However, a significant limitation remains: most existing risk fields rely on isotropic Gaussian distributions or simple distance metrics, which often fail to capture the asymmetric and directional nature of risk in high-speed, dynamic lane-changing scenarios.

\subsection{Optimization-Based Planning and Convexification}
Optimization-based methods have become the dominant paradigm for trajectory generation due to their rigorous handling of physical constraints. The primary challenge in this domain is the non-convex nature of collision avoidance constraints, which compromises real-time solvability. Early works utilized Mixed-Integer Programming (MIP) to model discrete decision logic, but its combinatorial complexity is often prohibitive for millisecond-level control loops.

Consequently, recent advancements have shifted towards "Convexification" techniques to ensure computational tractability. Deng et al.~\cite{deng2023safe} developed a spatial decomposition method that extracts convex corridors from the free space, enabling the use of efficient Quadratic Programming (QP) solvers. Li et al.~\cite{li2024cooperative} extended this concept to cooperative maneuvering, utilizing separating hyperplanes to guarantee safety margins among connected vehicles. In parallel, numerical solvers like the Iterative Linear Quadratic Regulator (iLQR) have gained traction for their rapid convergence in high-dimensional nonlinear systems. Recent studies by Chen et al.~\cite{chen2023integrated} and Li et al.~\cite{li2023efficient} demonstrated that combining iLQR with Control Barrier Functions (CBF) can achieve high-frequency planning while strictly enforcing safety bounds. 
Despite these successes, standard optimization frameworks often treat obstacles as hard constraints or static blocks. They frequently lack a unified, differentiable mechanism to dynamically weight soft risks (e.g., comfort vs. proximity) alongside hard safety constraints in complex interactive environments.

\subsection{Learning-Based and Hybrid Approaches}
With the proliferation of large-scale driving datasets, learning-based methods—specifically Reinforcement Learning (RL) and Imitation Learning (IL)—have emerged as powerful alternatives. RL agents can learn complex policies through trial-and-error, theoretically effectively handling the stochasticity of mixed traffic. Shi et al.~\cite{shi2024reinforcement} proposed a hierarchical RL framework to decouple high-level decision-making from low-level control, while Transformer-based architectures have been applied to capture long-term dependencies in human intent prediction~\cite{ xu2023transformer}.

However, pure learning-based methods face significant challenges regarding interpretability and worst-case safety guarantees (the "black-box" problem). In distributionally shifted scenarios, end-to-end networks may output hazardous actions~\cite{Ladosz2024}. To mitigate this, hybrid approaches combining learning with model-based safety layers are becoming increasingly popular. Zhu et al.~\cite{zhu2023bi} introduced a bi-level framework where an RL agent handles discrete behavioral decisions, leaving kinematic feasibility to a low-level optimizer. Similarly, Cheng et al.~\cite{cheng2023safe} utilized Model Predictive Shielding to override unsafe neural network actions. While promising, these hybrid methods often suffer from "switching instability" or conflicting objectives between the learning and planning modules. 


\section{Dynamic Risk Field and Convex Space}
This section introduces our integrated approach for trajectory planning by combining DRF-based collision risk assessment with convex feasible space generation. The framework ensures safety through continuous risk evaluation and kinematic feasibility through convex space constraints.

\subsection{Risk Field Modeling}
To quantify the potential collision risks during lane changes, a DRF is introduced to evaluate the interaction between host vehicle (HV) and surrounding vehicles (SVs). The total risk associated with each $SV_i$ is computed by
\begin{equation}
\label{drf}
R_i(x,y) = (R_{s,i}(x,y) + R_{d,i}(x,y)) F_e(x,y)
\end{equation}
where $R_{s,i}(x,y)$ and $R_{d,i}(x,y)$ are the static and dynamic risk factors, respectively.  $F_e(x,y)$ is an exponential decay factor. These three terms are detailed below. 

\begin{figure}[t]
    \centering
    \includegraphics[width=\linewidth]{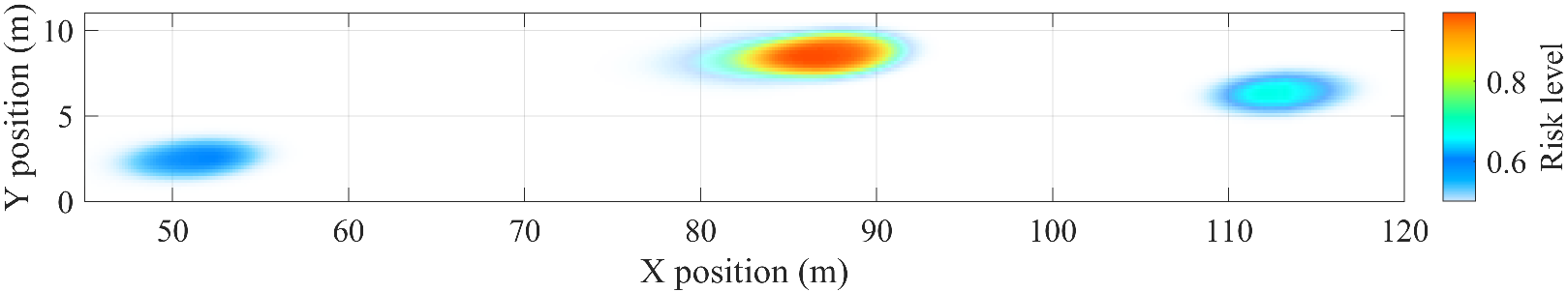}
    \vspace{-5mm}
\caption{Illustration of the dynamic risk field.}
    \label{drf_show}
\end{figure}

The static risk component $R_{s,i}(x,y)$ captures the spatial relationship between vehicles and is formulated using an obstacle-centered modified Gaussian distribution (see Fig. \ref{drf_show}):
\begin{equation}
\label{staticrisk}
R_{s,i}(x,y) = A_s \exp\left(-\left(\left(\frac{\Delta x'}{\sigma_x}\right)^2 + \left(\frac{\Delta y'}{\sigma_y}\right)^2\right)^\beta\right)
\end{equation}
where $A_s$ denotes the peak risk amplitude, while $\sigma_x$ and $\sigma_y$ define the risk distribution along longitudinal and lateral axes respectively, accounting for vehicle dimensions. The parameter $\beta$ determines how quickly the risk diminishes with distance. The coordinates $(\Delta x', \Delta y')$ are obtained by transforming the relative position to the obstacle's reference frame using its heading angle $\theta_i$, as follows:
\begin{equation}
\begin{bmatrix} \Delta x' \\ \Delta y' \end{bmatrix} = 
\begin{bmatrix} \cos\theta_i & \sin\theta_i \\ -\sin\theta_i & \cos\theta_i \end{bmatrix}
\begin{bmatrix} x - x_i \\ y - y_i \end{bmatrix}. \notag
\end{equation}

The dynamic risk term $R_{d,i}(x,y)$ accounts for the velocity-dependent interactions between vehicles. This component considers the relative velocity between the HV and $SV_i$, defined as $v_\text{rel,i} = v_{SVi} - v_\text{host}$, where $v_\text{host}$ is the velocity of the HV and $v_{SVi}$ is the velocity of $SV_i$:
\begin{equation}
\label{dynarisk}
R_{d,i}(x,y) \!=\!\! \frac{A_d \exp\left(-\frac{\Delta x'^2}{\sigma_v^2} - \frac{\Delta y'^2}{\sigma_y^2}\right)}{1 \!+\! \exp(-\text{sgn}(v_\text{rel})(\Delta x' \!-\! \alpha\sigma_x\text{sgn}(v_\text{rel})))}
\end{equation}
where $A_d$ scales the dynamic risk intensity, $\sigma_v = k_v|v_\text{rel}|$ adapts the risk distribution based on the relative velocity, and $\alpha$ modulates the position-dependent risk profile. 

The exponential decay factor $F_e(x,y)$ modifies the overall risk based on the separation distance between the evaluated position $(x,y)$ and the HV's current position $(x_\text{host}, y_\text{host})$:
\begin{equation}
\label{infulencefactor}
F_e(x,y) = \exp\left(-\sqrt{(x-x_\text{host})^2 + (y-y_\text{host})^2} / d_e\right)
\end{equation}
where $d_e$ is a characteristic length parameter that determines how quickly the risk decays with distance from the HV.

\subsection{Dynamic Convex Space Generation}
This subsection describes a rigorous approach for generating time-varying convex feasible spaces under vehicle kinematic and safety requirements. The approach integrates a growth-based expansion strategy with collision avoidance guarantees. The vehicle state is denoted by \(\mathbf{s} = [x,\, y,\, \theta,\, v,\, \phi]^\top \in \mathcal{X} \subset \mathbb{R}^{5}\), where \(x\) and \(y\) are planar coordinates, \(\theta\) is the heading angle, \(v\) is the longitudinal velocity, and \(\phi\) is the steering angle.


\begin{definition}[Dynamic Convex Feasible Space]
\label{def:dcfs}
Let \(\mathbf{p} = [p_x,\, p_y]^\top \in \mathbb{R}^2\) denote positions in plane. For state \(\mathbf{s}(t)\) at time \(t\), the rectangular time-varying convex feasible space \(\mathcal{C}(\mathbf{s},t)\) is defined as
\begin{equation}
\mathcal{C}(\mathbf{s},t) \!= \{\mathbf{p} \in \mathbb{R}^2 \mid \mathbf{A}(\mathbf{s},t)\mathbf{p} \leq \mathbf{b}(\mathbf{s},t), \Phi(\mathbf{p},\mathbf{s},t) \leq \mathbf{0}\}
\end{equation}
where the linear constraints \(\mathbf{A}(\mathbf{s},t)\mathbf{p} \leq \mathbf{b}(\mathbf{s},t)\) define the rectangular bounds:
\begin{equation}
\begin{bmatrix}
1 & 0 \\
-1 & 0 \\
0 & 1 \\
0 & -1
\end{bmatrix}
\begin{bmatrix}
p_x \\
p_y
\end{bmatrix}
\leq
\begin{bmatrix}
x_{\text{upper}}(t) \\
-x_{\text{lower}}(t) \\
y_{\text{upper}}(t) \\
-y_{\text{lower}}(t)
\end{bmatrix}.
\end{equation}
The bounds are iteratively expanded from the vehicle's current position with safety margins. The nonlinear constraints \(\Phi(\mathbf{p},\mathbf{s},t) \leq \mathbf{0}\) ensure collision avoidance with SVs with
\begin{equation}
\Phi_i(\mathbf{p},\mathbf{s},t) = \begin{cases}
1 & \text{if } \mathbf{p} \in \mathcal{O}_i(t) \\
0 & \text{otherwise}
\end{cases}, \quad i = 1,\ldots,N_{\text{obs}}
\end{equation}
where \(\mathcal{O}_i(t)\) represents the occupied region of the $SV_i$.

\end{definition}

\begin{figure}[t]
    \centering
    \includegraphics[width=0.95\linewidth]{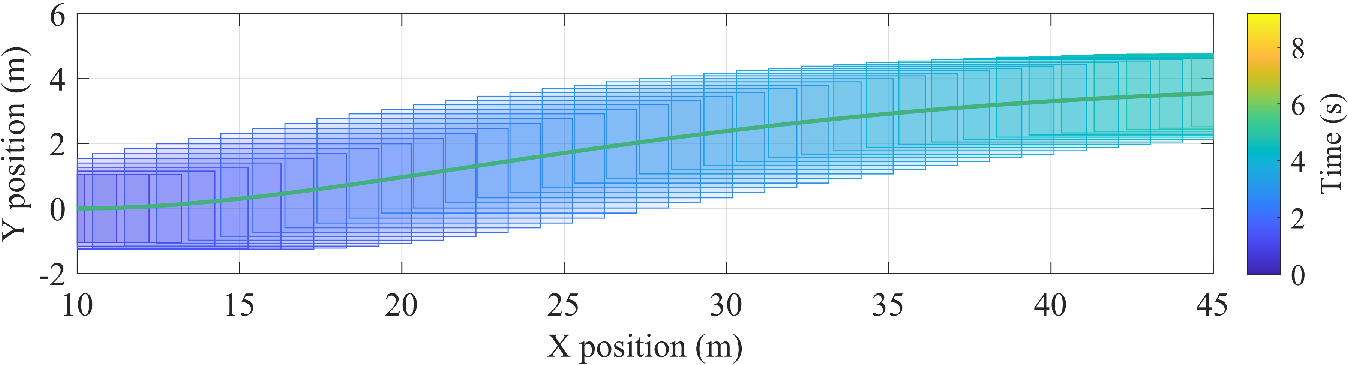}
    \vspace{-1mm}    
\caption{Evolution of the dynamic convex feasible space $\mathcal{C}(\mathbf{s},t)$ over time. The rectangles represent the growing feasible regions that satisfy both kinematic constraints and safety requirements, while the green curve represents a trajectory.}
    \label{fig:convex_space}
\end{figure}

Fig.~\ref{fig:convex_space} shows the evolution of the convex feasible space \(\mathcal{C}(\mathbf{s},t)\) over time, represented by growing rectangular regions at different time instances. This evolution adapts to the vehicle state through a continuous-time growth model dependent on velocity and heading, ensuring collision-free expansion.

\begin{proposition}[Growth Tensor Representation]
\label{prop:growthtensor}
There is a state-dependent growth tensor \(\mathcal{G}(\mathbf{s},t) \in \mathbb{R}^{2 \times 2}\) such that the evolution of the convex set \(\mathcal{C}(\mathbf{s},t)\) is described by
\begin{equation}
\frac{d}{dt}\mathcal{C}(\mathbf{s},t) = \{\mathbf{p} + \mathcal{G}(\mathbf{s},t)\mathbf{n}(\mathbf{p}) \mid \mathbf{p} \in \partial\mathcal{C}(\mathbf{s},t)\}
\end{equation}
where \(\mathbf{n}(\mathbf{p})\) denotes the outward normal vector at position \(\mathbf{p}\) on the boundary \(\partial\mathcal{C}(\mathbf{s},t)\). The growth tensor \(\mathcal{G}(\mathbf{s},t)\) admits the following decomposition: \(\mathcal{G}(\mathbf{s},t) = \alpha(\mathbf{s}) \, \Lambda(\theta) \, \Gamma(t)\)
with 
\begin{align}
 \alpha(\mathbf{s}) &= \min\{\,1 + \eta\,{v}/{v_{\text{ref}}},\,\alpha_{\text{max}}\},
 \nonumber\\
 \Lambda(\theta) &= \begin{bmatrix}
|\cos(\theta)| & -\sin(\theta) \\
\sin(\theta) & |\cos(\theta)|
\end{bmatrix}, \nonumber\;
\Gamma(t) = \gamma_{0}\,e^{-\lambda t}\,\mathbf{I}_{2}, \nonumber
\end{align}
where $\alpha(\mathbf{s})$
is the velocity-adaptive scaling factor with constants \(\eta, \alpha_{\text{max}}>0\) and the reference velocity \(v_{\text{ref}}>0\), \(\Lambda(\theta)\) is a heading-dependent matrix, and \(\Gamma(t)\) is a time-decaying component that ensures bounded expansion, with constants \(\gamma_{0}, \lambda>0\) and a \(2\times2\) identity matrix \(\mathbf{I}_{2}\).
\end{proposition}

Since the vehicle kinematics further impose curvature and velocity limits on the boundary of \(\mathcal{C}(\mathbf{s},t)\), the steering angle \(\phi\) must satisfy \(|\phi|\leq \phi_{\max}\), and the associated curvature is bounded. This leads to Lemma \ref{lem:kinematicfeas}.
\begin{lemma}[Kinematic Feasibility Conditions]
\label{lem:kinematicfeas}
Let \(\partial \mathcal{C}(\mathbf{s},t)\) be the boundary of the feasible space \(\mathcal{C}(\mathbf{s},t)\).
If $\mathbf{p}$ lies on \(\partial \mathcal{C}(\mathbf{s},t)\), the following kinematic constraints are necessary and sufficient for ensuring space feasibility:
\begin{equation}
|\kappa(\mathbf{p})|\leq \tan(\phi_{\text{max}}) / L,
~
\|\dot{\mathbf{p}}\|\leq v_{\text{max}},
~
|\ddot{\theta}|\leq \omega_{\text{max}}
\end{equation}
where \(\kappa(\mathbf{p})\) is the path curvature at \(\mathbf{p}\), \(L\) is the wheelbase length, and \(v_{\text{max}}\), \(\omega_{\text{max}}\) are upper bounds on velocity and yaw rate, respectively.
\end{lemma}

Collision avoidance with dynamic obstacles \(\{\mathcal{O}_{i}(t)\}\) is ensured by a separating hyperplane condition. Let each \(\mathcal{O}_{i}(t)\) be a convex set. Safe separation requires that, for each obstacle, there exists a time-varying hyperplane that strictly separates \(\mathcal{C}(\mathbf{s},t)\) from \(\mathcal{O}_{i}(t)\).

\begin{theorem}[Dynamic Separating Hyperplane Safety]
\label{thm:separatinghyperplane}
Safe collision-free motion is guaranteed if and only if, for every obstacle \(\mathcal{O}_{i}(t)\), there exist \(\mathbf{n}_i(t)\in \mathbb{S}^1\) and \(\gamma_i(t)\in \mathbb{R}\) such that
\begin{align}
\mathbf{n}_i(t)^\top\mathbf{p} &\leq \gamma_i(t), ~~ \forall \mathbf{p}\in \mathcal{C}(\mathbf{s},t) \\
\mathbf{n}_i(t)^\top\mathbf{q} &\geq \gamma_i(t)+ \delta_{\text{safe}}, ~~ \forall \mathbf{q}\in \mathcal{O}_{i}(t)
\end{align}
where \(\delta_{\text{safe}}>0\) is the required safety margin and $\mathbb{S}^1 =: \{\mathbf{x} \in \mathbb{R}^2 \mid \|\mathbf{x}\| = 1\}$ is the unit-circle set for normal directions.
\end{theorem}

The dynamic convex space then grows iteratively based on the current boundary and its outward normal, respecting all active constraints. At each infinitesimal time step \(\mathrm{d}t\), the convex set is expanded subject to kinematics and obstacle avoidance until it reaches a constraint boundary.
\begin{theorem}[Existence and Evolution of Feasible Spaces]
\label{thm:existence}

Given an initial convex space \(\mathcal{C}(\mathbf{s},0)\), the growth equation has
a unique solution 
\begin{equation}
\mathcal{C}(\mathbf{s},t+\mathrm{d}t) = 
\{\mathbf{p} + \mathcal{G}(\mathbf{s},t)\,\mathrm{d}t\,\mathbf{n}(\mathbf{p})
\mid
\mathbf{p} \in \mathcal{C}(\mathbf{s},t)
\}
\end{equation}
where \(\mathcal{G}(\mathbf{s},t)\) is as in Proposition~\ref{prop:growthtensor}, and \(\mathbf{n}(\mathbf{p})\) is the outward normal vector on the boundary \(\partial\mathcal{C}(\mathbf{s},t)\). The resulting \(\mathcal{C}(\mathbf{s},t)\) remains convex and satisfies the kinematic and safety constraints in Lemma~\ref{lem:kinematicfeas} and Theorem~\ref{thm:separatinghyperplane} for all \(t>0\).
\end{theorem}

\begin{figure}[t]
    \centering
    \includegraphics[width=\linewidth]{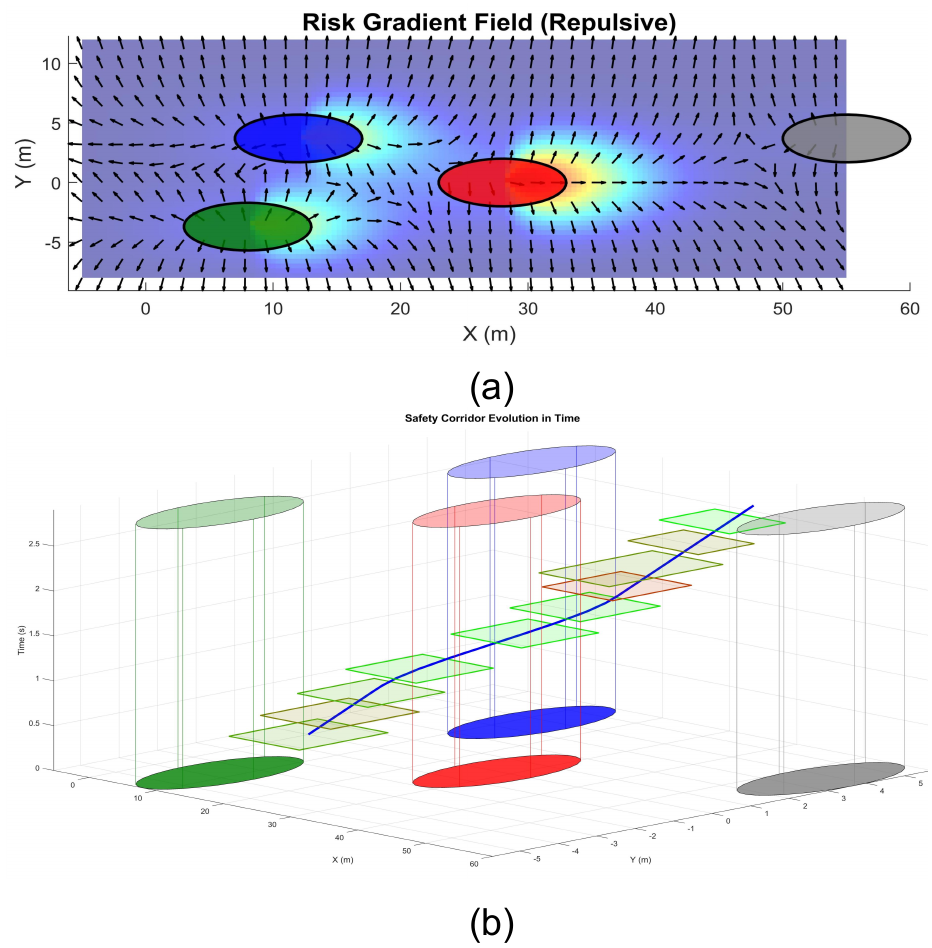}
    \caption{Risk gradient field (repulsive) and safety corridor evolution in time. 
    (a) Risk gradient field with repulsive vectors around obstacles. 
    (b) Time-evolving safety corridor along the planned trajectory.}
    \label{fig:risk_corridor}
\end{figure}

The analysis establishes a rigorous method to evolve time-varying convex sets that respect vehicle motion limits and ensure collision-free operation in dynamic environments. The tensor-based growth representation enables efficient computation while guaranteeing feasibility and safety.

Fig.~\ref{fig:risk_corridor} illustrates the relationship between the risk field and the resulting safety corridor.
As shown in Fig.~\ref{fig:risk_corridor}(a), each obstacle induces a repulsive risk potential whose gradient provides a spatially varying avoidance direction.
The superposition of multiple obstacles leads to a complex risk landscape, capturing both local proximity and geometric anisotropy effects.
Based on this risk representation, a time-varying safety corridor is constructed along the nominal trajectory, as depicted in Fig.~\ref{fig:risk_corridor}(b).
The corridor evolves over time and guarantees that the planned trajectory remains within a convex, collision-free feasible region at every planning step, thereby explicitly enforcing safety constraints during motion execution.

\section{Optimization-based Planning Framework}
Building upon the established DRF and convex space generation, we now formulate an optimization problem to generate feasible and safe lane chaning trajectories.
\subsection{Optimization Problem Formulation}
A finite-horizon optimal control problem is formulated to jointly address the DRF in \eqref{drf} and the time-varying convex feasible space from Definition~\ref{def:dcfs}. 


The discrete-time state is
$\mathbf{x}_{k} =[x_{k}, y_{k}, v_{k}, \theta_{k}, \phi_{k}, \dot{\phi}_{k}]^\top \in \mathbb{R}^6$
with longitudinal velocity \(v_{k}\), heading angle \(\theta_{k}\), steering angle \(\phi_{k}\), and steering rate \(\dot{\phi}_{k}\), where steering angle and its rate are considered to better handle the steering dynamics. The control input is 
$
\mathbf{u}_{k} 
\,=\,
[
a_{k},\;\ddot{\phi}_{k}
]^\top
$, 
where \(a_{k}\) is the longitudinal acceleration and \(\ddot{\phi}_{k}\) is the steering acceleration. The system evolves via
\begin{equation}\label{eq:system}
\mathbf{x}_{k+1} = f(\mathbf{x}_{k}, \mathbf{u}_{k})
\end{equation}
from the given the initial state \(\mathbf{x}_{0} = \mathbf{x}^{\mathrm{init}}\in\mathbb{R}^{6}\). The function \(f\) is obtained from the continuous-time kinematic bicycle model \cite{9658226} using Euler method with a sampling period \(\Delta t\). A receding-horizon strategy employs a planning horizon \(N>0\), so \(k=0,\dots,N-1\). The total horizon cost is
\begin{equation}
J\bigl(\{\mathbf{x}_{k}\}, \{\mathbf{u}_{k}\}\bigr)
\,=\,
\sum_{k=0}^{N-1}
\ell(\mathbf{x}_{k}, \mathbf{u}_{k})
\;+\;
\ell_{T}(\mathbf{x}_{N}).
\end{equation}
The stage cost $\ell(\mathbf{x}_{k}, \mathbf{u}_{k})$ penalizes the reference deviation, control effort, and collision risk, and it is defined as
\begin{equation}
\ell(\mathbf{x}_{k}, \mathbf{u}_{k})
=
(\mathbf{x}_{k}
-
\mathbf{x}_{k}^{\mathrm{ref}}
)^\top
\mathbf{Q}\,
(\mathbf{x}_{k}
-
\mathbf{x}_{k}^{\mathrm{ref}}
)
+
\mathbf{u}_{k}^\top\,\mathbf{R}\,\mathbf{u}_{k}
+
\gamma\,\mathcal{R}(\mathbf{x}_{k})
\notag
\end{equation}
where \(\{\mathbf{x}_{k}^{\mathrm{ref}}\}\) is the reference trajectory prescribing nominal poses and velocities, and \(\mathbf{Q}\), \(\mathbf{R}\) are positive semidefinite weighting matrices on the state and control. The term \(\mathcal{R}(\mathbf{x}_{k})\) aggregates the collision risk at time step \(k\) from \eqref{drf} with a risk weighting parameter \(\gamma>0\).

The terminal cost $\ell_{T}(\mathbf{x}_{N})$ ensures that \(\mathbf{x}_{N}\) approaches the desired terminal state 
and is defined as
\begin{equation}
\ell_{T}(\mathbf{x}_{N})
\,=\,
(\mathbf{x}_{N}
-
\mathbf{x}_{N}^{\mathrm{ref}}
)^\top
\mathbf{Q}_{T}\,
(\mathbf{x}_{N}
-
\mathbf{x}_{N}^{\mathrm{ref}}
)
\end{equation}
with the positive semidefinite weighting matrix \(\mathbf{Q}_{T}\). 

All states \(\mathbf{x}_{k}\) must remain within the time-varying convex set \(\mathcal{C}(\mathbf{s}_{k},t_{k})\) from Theorem~\ref{thm:existence} to guarantee collision-free, dynamically feasible motion. In discrete form, if \(\mathbf{x}_{k}=[\,x_{k},\,y_{k},\,v_{k},\,\theta_{k},\,\phi_{k},\,\dot{\phi}_{k}\,]^\top\), then 
$
[p_{x},\,p_{y}]^\top 
=
[x_{k},\,y_{k}]^\top
\in
\mathcal{C}\bigl(\mathbf{s}_{k}, t_{k}\bigr),
$
and 
$
\mathbf{s}_{k}
=
[x_{k},\, y_{k},\, \theta_{k},\, v_{k},\, \phi_{k},\, \dot{\phi}_{k}]^\top.
$
The optimization problem for trajectory planning is formulated as
\begin{equation}\label{eq:OP}
\begin{aligned}
&\min_{\{\mathbf{u}_{k}\}_{k=0}^{N-1}}
J \left(\{\mathbf{x}_{k}\}, \{\mathbf{u}_{k}\} \right)
\;\\
\mathrm{s.t.} \quad&
\;
\eqref{eq:system},~ [x_{k},\,y_{k}]^\top\in \mathcal{C}(\mathbf{s}_{k},t_{k}),
\;
|\phi_{k}|\le \phi_{\max}.
\end{aligned}
\end{equation}
The steering limit \(|\phi_{k}|\le\phi_{\max}\) enforces bounded curvature, while the convex region \(\mathcal{C}\) ensures collision avoidance per Theorem~\ref{thm:separatinghyperplane}. The risk function \(\mathcal{R}\) and convex constraints jointly optimize obstacle avoidance. For this nonlinear optimization problem \eqref{eq:OP}, an iLQR-based algorithm is proposed in Section \ref{subsec:iLQR} to enable efficient and safe lane changing.

\subsection{iLQR-based Solution}\label{subsec:iLQR}
The iLQR method is selected for solving the constrained nonlinear optimization problem \eqref{eq:OP} due to its real-time efficiency and ability to handle state and control constraints \cite{10452817}. By leveraging recursive value iteration, it efficiently enforces kinematic constraints and collision avoidance, making it ideal for autonomous driving applications.
A constrained iterative LQR algorithm solves the discrete-time problem by successively linearizing the dynamics and quadratically approximating the cost function. Let \(\{\mathbf{x}_{k}^{(0)}, \mathbf{u}_{k}^{(0)}\}\) be an initial guess of state and control over the horizon. At iteration \(\ell\), the vehicle dynamic system \eqref{eq:system} is linearized as
\begin{align}
\mathbf{x}_{k+1} 
\approx
f\bigl(\mathbf{x}_{k}^{(\ell)}, \mathbf{u}_{k}^{(\ell)}\bigr)
\!+\!
A_{k}^{(\ell)}
\delta\mathbf{x}_{k}
\!+\!
B_{k}^{(\ell)}
\delta\mathbf{u}_{k}
\end{align}
where 
$
A_{k}^{(\ell)}
=
\frac{\partial f}{\partial \mathbf{x}}
(\mathbf{x}_{k}^{(\ell)}, \mathbf{u}_{k}^{(\ell)})$
and
$
B_{k}^{(\ell)}
=
\frac{\partial f}{\partial \mathbf{u}}
(\mathbf{x}_{k}^{(\ell)}, \mathbf{u}_{k}^{(\ell)}). 
$ 
$\delta\mathbf{x}_{k} = \mathbf{x}_{k}-\mathbf{x}_{k}^{(\ell)}$ and $\delta\mathbf{u}_{k} = \mathbf{u}_{k}-\mathbf{u}_{k}^{(\ell)}$ are the state and control deviations, respectively.

The stage cost is approximated quadratically around the current trajectory as
\begin{align}
\ell(\mathbf{x}_{k}, \mathbf{u}_{k})
\approx\;&
\ell\bigl(\mathbf{x}_{k}^{(\ell)}, \mathbf{u}_{k}^{(\ell)}\bigr)
+
\nabla_{\mathbf{x}}\ell\bigl|_{(\ell)}^\top
\delta\mathbf{x}_{k}
+
\nabla_{\mathbf{u}}\ell\bigl|_{(\ell)}^\top
\delta\mathbf{u}_{k}
\notag\\
&+
\tfrac12
\begin{bmatrix}
\delta\mathbf{x}_{k}\\
\delta\mathbf{u}_{k}
\end{bmatrix}^\top
\begin{bmatrix}
\nabla_{\mathbf{xx}}^2\ell & \nabla_{\mathbf{xu}}^2\ell\\
\nabla_{\mathbf{ux}}^2\ell & \nabla_{\mathbf{uu}}^2\ell
\end{bmatrix}
\bigl|_{(\ell)}
\begin{bmatrix}
\delta\mathbf{x}_{k}\\
\delta\mathbf{u}_{k}
\end{bmatrix}
\end{align}
where $\nabla_{\mathbf{xx}}^2\ell$, $\nabla_{\mathbf{xu}}^2\ell$, $\nabla_{\mathbf{ux}}^2\ell$ and $\nabla_{\mathbf{uu}}^2\ell$ represent the second-order partial derivatives of the stage cost with respect to states and controls, \(|_{(\ell)}\) denotes evaluation at the current iteration point \((\mathbf{x}_{k}^{(\ell)}, \mathbf{u}_{k}^{(\ell)})\). The gradients \(\nabla_{\mathbf{x}}\ell\) and \(\nabla_{\mathbf{u}}\ell\) incorporate risk penalties and boundary constraints. 

\begin{figure}[t]
    \centering
    \includegraphics[width=0.7\linewidth]{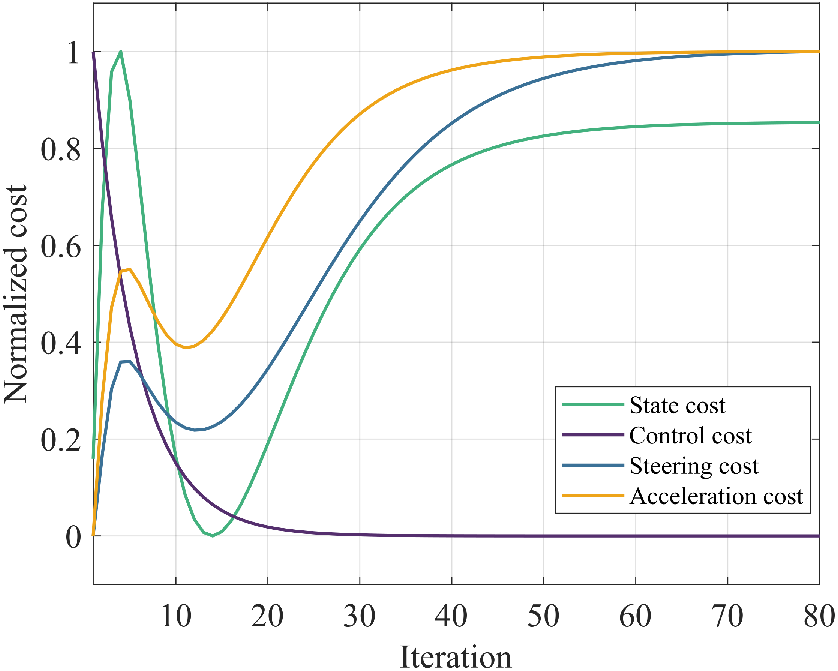}
    \vspace{-1mm}
    \caption{Convergence analysis of proposed method.}
    \label{fig:cost_convergence}
\end{figure}

\begin{figure*}[t]
    \centering
    \begin{subfigure}[t]{1\textwidth}
        \includegraphics[width=\textwidth]{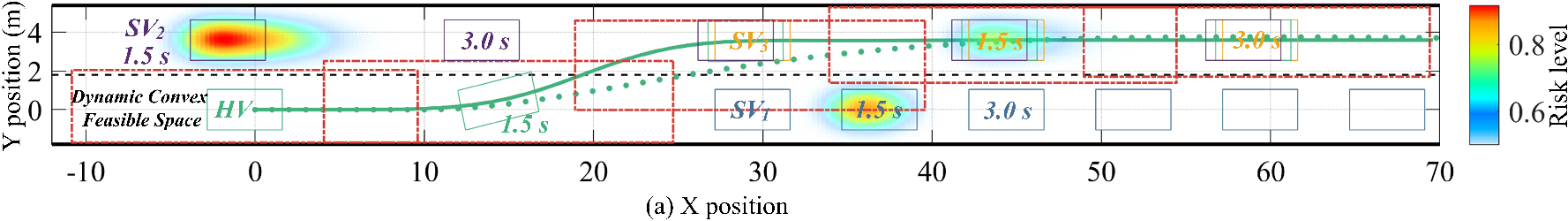}
        \vspace{-0.8cm}
        \label{fig9a}
    \end{subfigure}
    
    \vspace{0cm} 

    \begin{subfigure}[t]{0.24\textwidth}
        \includegraphics[width=\textwidth]{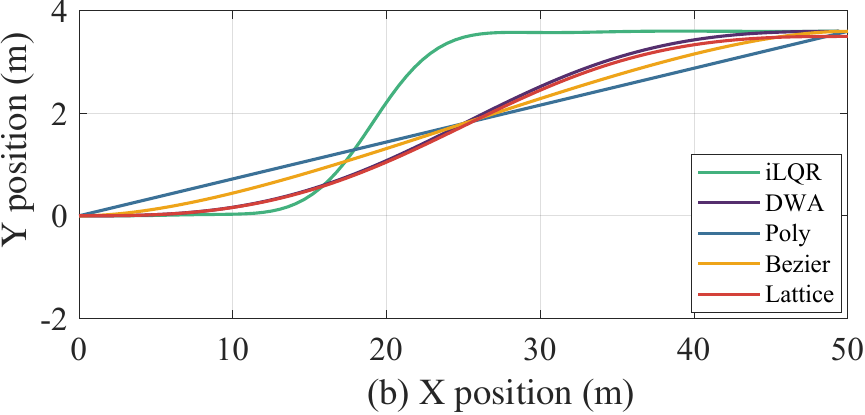}
        \vspace{-0.6cm}
        
        \label{fig9b}
    \end{subfigure}
    \begin{subfigure}[t]{0.24\textwidth}
        \includegraphics[width=\textwidth]{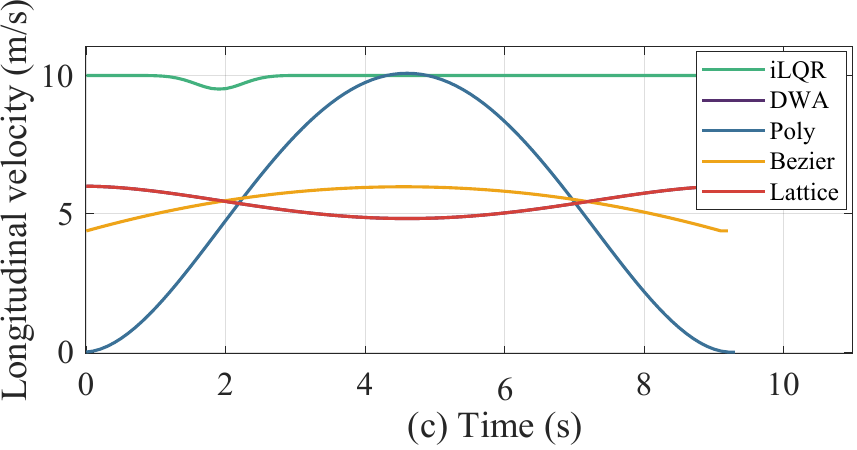}
        \vspace{-0.6cm}
        
        \label{fig9c}
    \end{subfigure}
    \begin{subfigure}[t]{0.24\textwidth}
        \includegraphics[width=\textwidth]{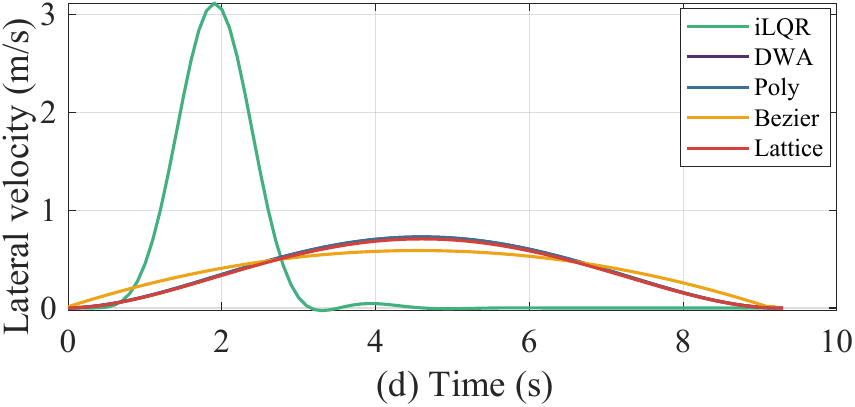}
        \vspace{-0.6cm}
        
        \label{fig9d}
    \end{subfigure}
    \begin{subfigure}[t]{0.24\textwidth}
        \includegraphics[width=\textwidth]{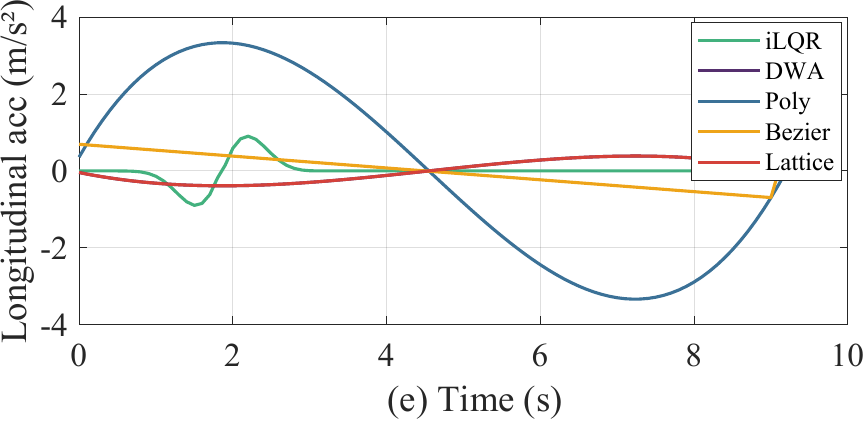}
        \vspace{-0.6cm}
        
        \label{fig9e}
    \end{subfigure}
    \vspace{-2mm}
\caption{Comparison of different lane changing trajectory planning methods. (a) Visualization of dynamic convex feasible space as red rectangles and DRF at \(t = 1.5~\mathrm{s}\). The green line shows the planned trajectory, green dots represent the trajectory before optimization. (b) Comparison of lane changing using different trajectory planning methods. (c) Comparison of longitudinal velocities. (d) Comparison of lateral velocities. (e) Comparison of longitudinal accelerations.}
    \label{fig4}
\end{figure*}

The backward pass computes local feedback \(\mathbf{K}_{k}\) and feedforward \(\mathbf{k}_{k}\) gains by propagating the second-order cost-to-go, subject to the risk and convex feasibility constraints. Define \(V_{x}\in\mathbb{R}^n\) and \(V_{xx}\in\mathbb{R}^{n\times n}\) as the gradient and Hessian of the value function, respectively, where \(n\) is the dimension of the state space, then
\begin{align}
\mathbf{K}_{k}
&=
-\,Q_{uu}^{-1}\,Q_{ux},
\quad
\mathbf{k}_{k}
=
-\,Q_{uu}^{-1}\,Q_{u}
\end{align}
where
\begin{align}
Q_{u}
&=
B_{k}^{(\ell)T}\,V_{x}
+
\nabla_{\mathbf{u}}\ell\bigl|_{(\ell)},
\quad
Q_{x}
=
A_{k}^{(\ell)T}\,V_{x}
+
\nabla_{\mathbf{x}}\ell\bigl|_{(\ell)},
\nonumber\\
Q_{uu}
&=
\mathbf{R}
+
B_{k}^{(\ell)T}\,V_{xx}\,B_{k}^{(\ell)},
\quad
Q_{ux}
=
B_{k}^{(\ell)T}\,V_{xx}\,A_{k}^{(\ell)},
\nonumber\\
Q_{xx}
&=
\mathbf{Q}
+
A_{k}^{(\ell)T}\,V_{xx}\,A_{k}^{(\ell)}. \nonumber
\end{align}
The updated cost-to-go satisfies
$
V_{x}
=
Q_{x}
-
\mathbf{K}_{k}^\top\,
Q_{uu}\,\mathbf{k}_{k}$,
and 
$
V_{xx}
=
Q_{xx}
-
\mathbf{K}_{k}^\top\,Q_{uu}\,\mathbf{K}_{k}.
$

The forward pass updates the control action by
\begin{equation}
\mathbf{u}_{k}^{(\ell+1)}
=
\mathbf{u}_{k}^{(\ell)}
+
\alpha\,\mathbf{k}_{k}
+
\mathbf{K}_{k}\,
(\mathbf{x}_{k}^{(\ell+1)}-\mathbf{x}_{k}^{(\ell)})
\end{equation}
where \(\alpha\in(0,1]\) is a line search parameter. 

Each forward step checks feasibility \(\mathbf{p}_{k}=[x_k,y_k]^\top\in \mathcal{C}(\mathbf{s}_{k},t_{k})\) for the position components; if violated, the algorithm adjusts gradients or re-projects to remain strictly in \(\mathcal{C}\). Risk gradients from \(\mathcal{R}(\mathbf{x}_{k})\) are incorporated to discourage motion into hazardous zones. This backward-forward iteration continues until convergence or maximum iterations. The optimization process operates at \(10~\mathrm{Hz}\) with \(0.1~\mathrm{s}\) discretization, with trajectories typically maintaining validity for \(0.3\text{--}0.5~\mathrm{s}\) and re-optimization frequency adaptable to environmental changes.

The final solution \(\{\mathbf{x}_{k}^{(\ast)}, \mathbf{u}_{k}^{(\ast)}\}\) satisfies both kinematic feasibility and collision avoidance while minimizing \(\gamma\,\mathcal{R}(\mathbf{x}_{k})\). By unifying potential-field risk modeling and convex set growth in a single receding-horizon framework, real-time implementations can generate trajectories that adapt to dynamic obstacles and continuously remain safe driving.

\subsection{Stability Analysis}
The risk-sensitive planner must remain within \(\mathcal{C}\) while penalizing risk. An iterative procedure refines an initial nominal solution until a local minimum of the objective \(J\) is found. The risk penalty \(\gamma\,\mathcal{R}(\mathbf{x}_{k})\) shifts the trajectory away from high-risk regions, while the convex boundary constraint enforces safe separation from obstacles and road edges.

To analyze the algorithm's convergence, Fig.~\ref{fig:cost_convergence} shows the evolution of normalized cost components including state deviation, control effort, steering, and acceleration costs. The stabilization of these components around iteration \(60-70\) indicates convergence to a locally optimal solution that satisfies both feasibility (\(\mathbf{x}_{k}\in\mathcal{C}\)) and safety (i.e., low \(\gamma\,\mathcal{R}(\mathbf{x}_{k})\)). This demonstrates that our iLQR process effectively balances risk minimization with constraint satisfaction, consistently generating stable and feasible trajectories.
\begin{figure}[t] 
    \centering
  \includegraphics[width=1\linewidth]{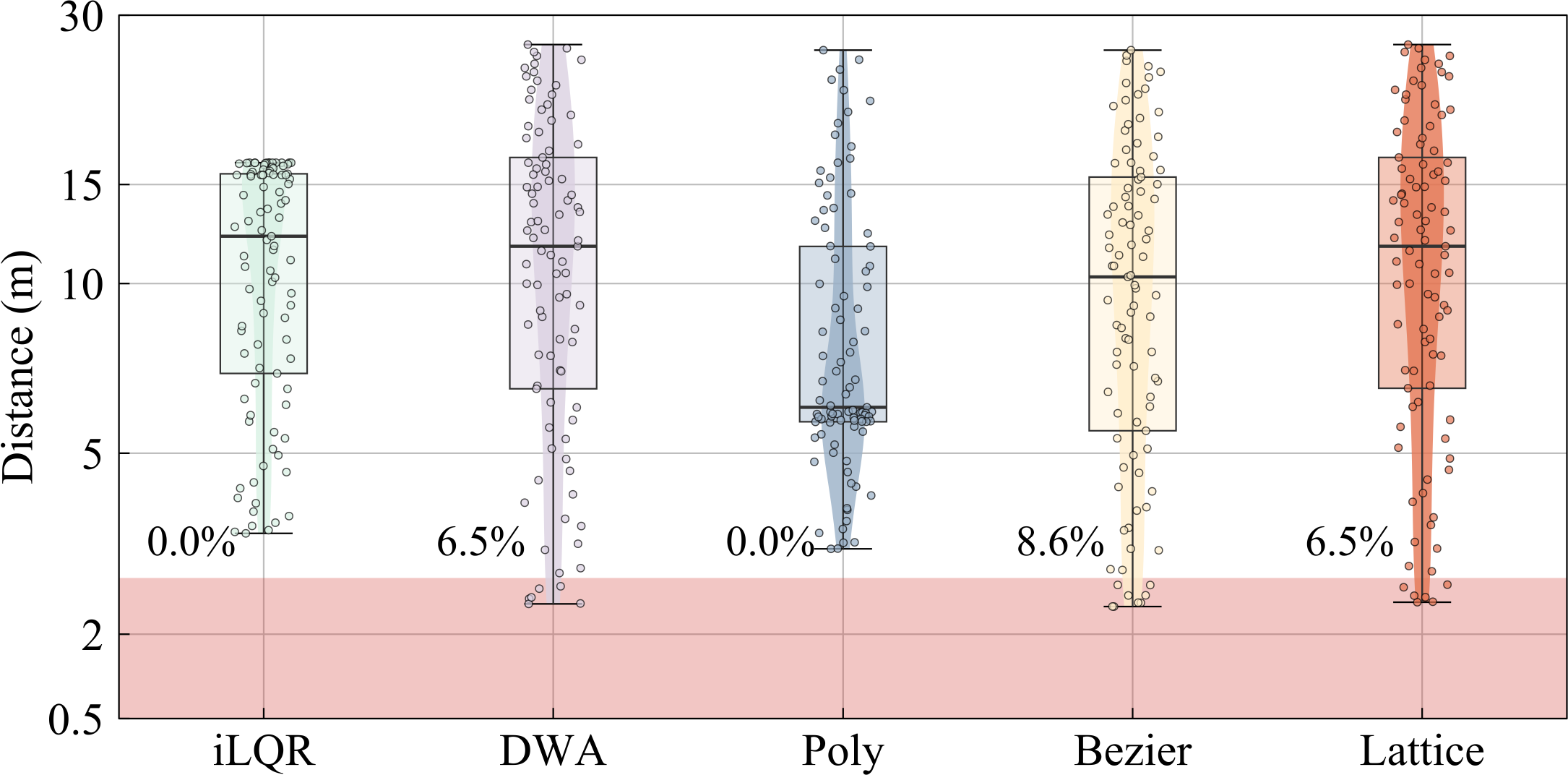}
    \vspace{-6mm}
    \caption{ Comparison of minimum distances to SVs for different planning methods. The red shaded area indicates unsafe.}
    \label{fig5}
\end{figure}

\begin{figure*}[t]
    \centering
    \includegraphics[width=\textwidth]{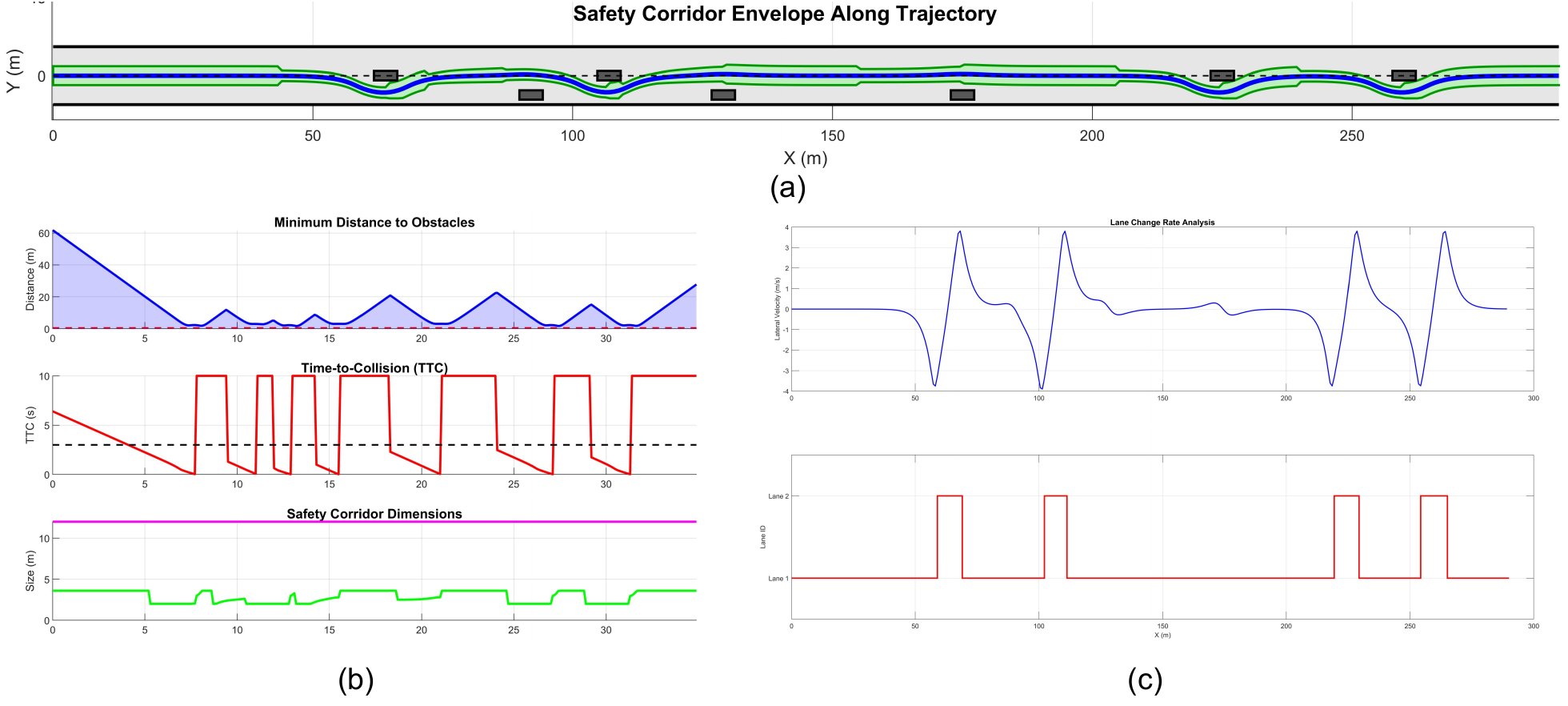}
    \caption{Safety corridor envelope and safety metrics along the planned lane-change trajectory in a two-lane road scenario.
    (a) Safety corridor envelope along the trajectory: the planned path (blue) remains inside the corridor bounds (green) under multi-vehicle constraints (gray blocks).
    (b) Safety indicators along the maneuver, including the minimum clearance to obstacles, the time-to-collision (TTC), and the corridor dimension evolution; dashed reference lines denote typical safety thresholds.
    (c) Lane-change dynamics: the lane-change rate peaks coincide with lane ID transitions, confirming consistent detection of lane-change events.}
    \label{fig:safety_corridor_metrics}
\end{figure*}

\section{EXPERIMENTAL EVALUATION}
\label{sec6}
Simulations are conducted in MATLAB 2024b to evaluate the proposed approach for safe and efficient autonomous driving in complex lane-changing scenarios. We compare our iLQR-based method with several baseline approaches including Dynamic Window Approach (DWA)~\cite{Dobrevski2024}, polynomial-based planning (Poly)~\cite{9749979}, Bezier curve planning~\cite{Liu20244}, and lattice-based planning~\cite{Xia20244}.

\begin{table}[t]
\centering
\setlength{\tabcolsep}{5pt}
\captionsetup{
    labelfont={sc}, 
    textfont={sc}, 
    labelsep=colon, 
    skip=1em,       
    singlelinecheck=false,
    justification=centering,
    format=plain
}
\caption{\protect\\ Comparison of Algorithm Performances}
\vspace{-2mm}
\label{tab1}
\begin{threeparttable} 
\begin{tabular}{c|c|c|c|c|c}
\hline \hline
\multirow{2}{*}{Methods} & $v_x$ & $v_y$ & $\text{j}_x$ & $\mathrm{D}$   & $\mathrm{T}$\\
 & (m/s$^2$) & (m/s$^2$) & (m/s$^3$) &(m) & (s)  \\
\hline
iLQR & $\mathbf{9.96}$  & $\mathbf{0.40}$   & $0.38$  & $\mathbf{28.59}$  & $\mathbf{2.84}  \pm {\mathbf{0.03}}$ \\
\hline
DWA &$5.38$  & $0.39$   & $\mathbf{0.15}$  & $45.82$  & $8.04 \pm {0.04}$ \\
\hline
Ploy  &$5.32$  & $0.38$   & $1.38$  & $57.62$  &  $11.08 \pm {0.05}$\\
\hline
Bezier & $5.42$  & $0.39$   & $0.22$  & $49.11$  & $8.63\pm {0.05}$ \\
\hline
Lattice& $5.38$  & $0.37$   & $0.16$  & $42.32$  & $7.29\pm {0.05}$ \\
\hline\hline
\end{tabular}
\begin{tablenotes}
\small
\item$v_x$, $v_y$: average longitudinal and lateral velocities; $\text{j}_x$: average longitudinal jerks; $\mathrm{D}$: average lane changing distance; $\mathrm{T}$: average lane changing time.
\end{tablenotes}
\end{threeparttable}
\vspace{-3 mm}
\end{table}

The comparative analysis demonstrates several key advantages of our method:

\begin{figure*}[t]
    \centering
    \includegraphics[width=0.95\linewidth]{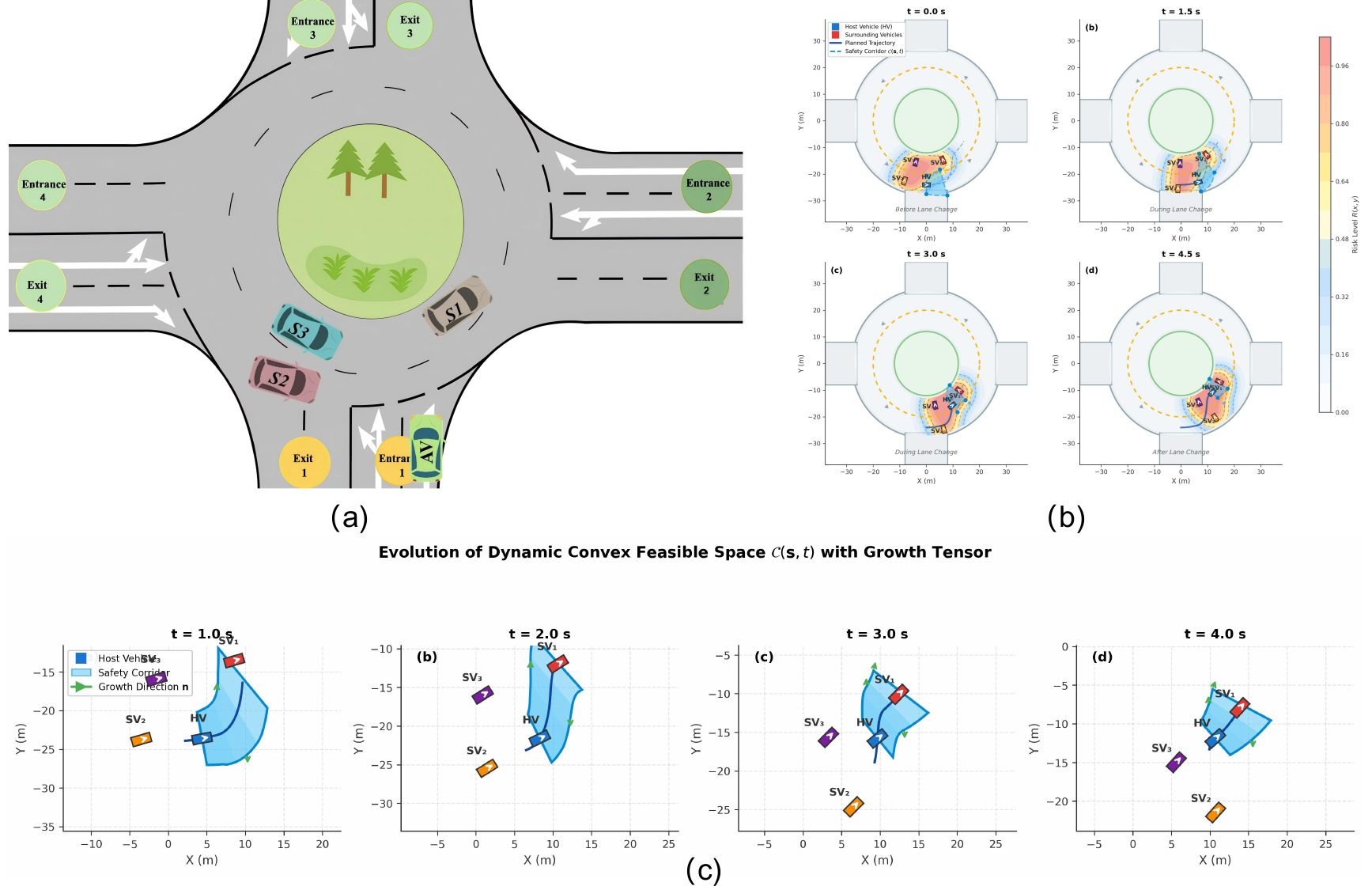}
    \vspace{-2mm}
    \caption{
    Overall visualization of the proposed DRF-iLQR planning pipeline in a dual-lane roundabout.
    (a) Scenario layout.
    (b) DRF evolution.
    (c) Evolution of the Dynamic Convex Feasible Space $\mathcal{C}(\mathbf{s},t)$ under the Growth Tensor.
    The green trajectory demonstrates that the HV maintains a safe, feasible, and risk-aware motion while interacting with multiple surrounding vehicles.
    }
    \label{fig:roundabout_overview}
\end{figure*}
1) \textbf{Trajectory Efficiency}: As shown in Fig.~\ref{fig4}(a) and Fig.~\ref{fig4}(b), our iLQR-based method achieves significantly shorter lane-change distances, completing the maneuver within \(28.59~ \mathrm{m}\) compared to \(42-57~\mathrm{m}\) required by other approaches. The optimized trajectory (green line) exhibits smooth path characteristics while maintaining consistency with the dynamic convex feasible space constraints.

2) \textbf{Velocity Control}: The longitudinal velocity profile in Fig.~\ref{fig4}(c) demonstrates our iLQR superior speed regulation capabilities, maintaining a stable velocity profile around \(10~\mathrm{m/s}\) compared to the fluctuating patterns of other approaches.

3) \textbf{Lateral Motion Strategy}: Fig.~\ref{fig4}(d) highlights our method's aggressive yet efficient lane-changing strategy. The initial lateral velocity peak reflects an intentional trade-off of motion smoothness for significant time efficiency, enabling a rapid lane change in \(2.84 ~\mathrm{s}\)  without collisions—much faster than the \(7-11 ~\mathrm{s}\) of other approaches. Despite higher instantaneous lateral velocity, the controller swiftly stabilizes the motion, balancing rapid maneuvering with vehicle stability. This aligns with our optimization objective of minimizing lane-changing time while ensuring safe operation.

4) \textbf{Acceleration Management}: The longitudinal acceleration profile in Fig.~\ref{fig4}(e) shows that our iLQR method maintains smoother acceleration changes within a reasonable range of \(\pm 1~\mathrm{m/s^2}\), whereas other methods exhibit more aggressive acceleration patterns.

These results validate that our approach successfully integrates risk awareness with convex feasibility constraints, producing trajectories that are not only dynamically feasible but also optimized for achieving safety and comfort simultaneously. The iterative nature of our iLQR solution effectively refines the initial trajectory (green dots) into a smooth, executable path (green line) that satisfies all constraints.

\begin{figure*}[!t]
    \centering
    \includegraphics[width=\linewidth]{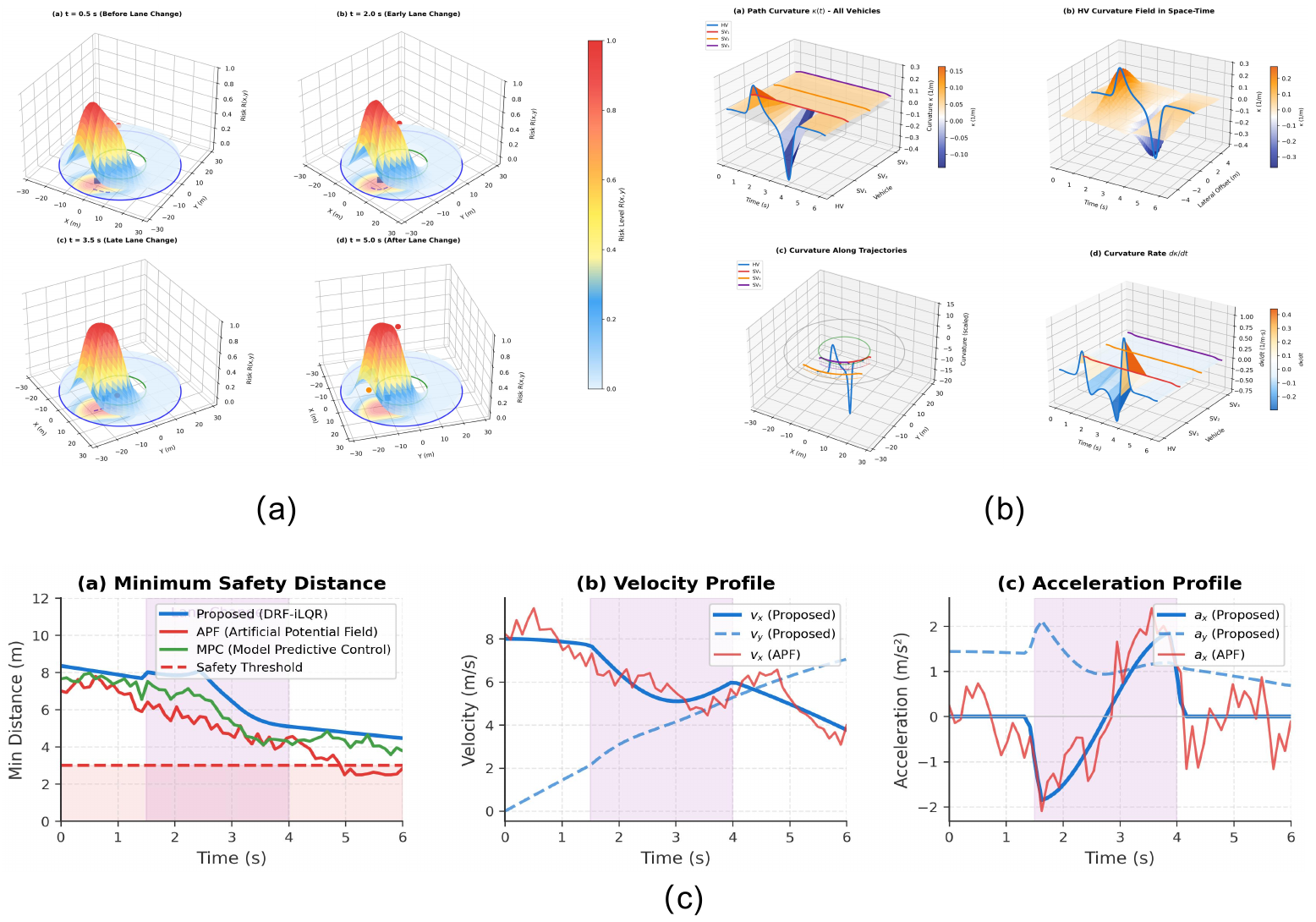}
    \vspace{-2mm}
    \caption{
    Quantitative analysis of the proposed DRF-iLQR planner.
    (a) Evolution of the DRF along the lane-change manoeuvre.
    (b) Path curvature and curvature-rate profiles for all vehicles and the HV.
    (c) Comparison of minimum safety distance, velocity, and acceleration profiles between the proposed method, APF, and MPC baselines.
    }
    \label{fig:risk_curv_safety}
\end{figure*}


To evaluate the safety performance of different planning methods, we analyze the minimum distance between the HV and SVS. Fig.~\ref{fig5} compares these distances, highlighting unsafe zones below \(3 \mathrm{m}\) (red shading). Our iLQR-based method achieves a \(0.0\%\) collision rate, matching the polynomial method, while DWA, Bezier, and lattice-based methods have collision rates of \(6.5\%\), \(8.6\%\), and \(6.5\%\), respectively. With a median distance of \(\sim10~\mathrm{m}\) and consistent safety margins, our method balances safety and performance, avoiding excessive conservatism while efficiently utilizing space.

Table~\ref{tab1} compares key performance metrics of various planning methods. Our iLQR-based approach achieves the highest longitudinal velocity (\(9.96~\mathrm{m/s^2}\)) and comparable lateral velocity (\(0.40~\mathrm{m/s^2}\)). It also achieves the shortest lane-changing distance (\(28.59~\mathrm{m}\)) and time (\(2.84~\mathrm{s}\)) with minimal variation (\(\pm~ 0.03~\mathrm{s}\)), demonstrating efficiency and consistency. Although its longitudinal jerk (\(0.38~\mathrm{m/s^3}\)) is slightly higher than DWA, this trade-off ensures a balance between smoothness and efficiency.

Fig.~\ref{fig:safety_corridor_metrics} reports the corridor-level safety verification for the paper-style lane-change demo.
The road length is set to 300\,m with two lanes of width 3.7\,m, whose centerlines are defined at $y=0$ (Lane~1) and $y=-3.7$ (Lane~2).
The ego vehicle follows a discrete-time trajectory with $\Delta t=0.1$\,s and a target speed of 15\,m/s.
The planning horizon is computed as $N=\mathrm{round}(L/v/\Delta t)+50$, resulting in approximately 250 steps for $L=300$\,m.
The initial trajectory is a constant-speed lane-following motion ($y=0$), which is then refined by an iterative potential-field update that combines obstacle-induced repulsion, lane-center attraction, and a smoothness regularizer.

As shown in Fig.~\ref{fig:safety_corridor_metrics}(a), an obstacle-aware safety corridor is generated along the optimized trajectory, and the planned path remains fully enclosed by the corridor envelope, indicating constraint-consistent motion throughout the horizon.
Fig.~\ref{fig:safety_corridor_metrics}(b) further quantifies the safety evolution: the minimum clearance decreases when the ego vehicle approaches surrounding traffic and recovers after passing; the TTC curve exhibits low values in critical segments and is capped at 10\,s for visualization, with a warning threshold at 3\,s.
The corridor dimensions shrink near dense obstacles and expand in free space, reflecting the adaptive feasibility region induced by the environment.
Finally, Fig.~\ref{fig:safety_corridor_metrics}(c) shows that the peaks of lane-change rate align with discrete lane ID transitions, confirming that the lane-change events are temporally and spatially consistent with the trajectory evolution.

Fig.~\ref{fig:roundabout_overview}(a) shows the dual-lane roundabout with four entrances and four exits. The host vehicle (HV) enters at Entrance~1 and aims to perform a lane change while interacting with multiple surrounding vehicles (SVs).  
The scenario intentionally includes heterogeneous SV behaviours (slower leading vehicle, aggressive merging vehicle, and a fast rear vehicle) to stress-test the proposed dynamic planning framework.

\subsection{Evolution of Dynamic Risk Field}
Fig.~\ref{fig:roundabout_overview}(b) presents the time evolution of the DRF at $t=0$, $1.5$, $3.0$, and $4.5~\text{s}$.  
At $t=0$, the risk distribution is nearly isotropic because the HV has just entered the roundabout and no close interaction is occurring.  
As time progresses to $t=1.5$ and $3.0~\text{s}$, a strong asymmetric risk lobe emerges in front of SV\textsubscript{0} and SV\textsubscript{1}, consistent with the dynamic risk model in~\eqref{dynarisk} where the relative velocity and relative heading increase the collision likelihood.  
Near $t=3.0$--$4.5~\text{s}$, the DRF exhibits a strong lateral gradient as SV\textsubscript{3} accelerates toward the inner lane.  
This behaviour demonstrates the DRF's ability to continuously quantify the \emph{directional and velocity-dependent collision risk}, enabling gradient-based avoidance within the iLQR optimizer.

\subsection{Evolution of Convex Feasible Space}
Fig.~\ref{fig:roundabout_overview}(c) shows the expansion and deformation of the dynamic convex feasible space $\mathcal{C}(\mathbf{s},t)$ from $t=1.0$ to $4.0~\text{s}$.  
At $t=1.0~\text{s}$, the feasible set expands primarily forward due to the velocity-dependent scaling factor $\alpha(\mathbf{s})$ in Proposition~\ref{prop:growthtensor}.  
Because no obstacle is yet constraining the HV laterally, the set remains symmetrically shaped around the vehicle.  

By $t=2.0~\text{s}$, SV\textsubscript{1} has moved closer, and the separating hyperplane condition in Theorem~\ref{thm:separatinghyperplane} trims the right-hand side of $\mathcal{C}(\mathbf{s},t)$, preventing any unsafe expansion toward the conflict zone.

At $t=3.0~\text{s}$ and $4.0~\text{s}$, the feasible set becomes increasingly skewed toward the left side as both SV\textsubscript{0} and SV\textsubscript{3} create high-risk zones.  
The Growth Tensor naturally rotates the feasible region according to the HV's heading direction via $\Lambda(\theta)$, while the exponential decay term $\Gamma(t)$ prevents unbounded expansion.  
These results confirm that the convex feasible space continuously adapts to traffic geometry, vehicle kinematics, and dynamic obstacles.

\subsection{Safety-Aware Motion Behaviour}
Combining the DRF with the convex feasible region ensures that the iLQR-generated trajectory (green curve in the figures) always remains inside the safe, dynamically reachable area:  
\begin{itemize}
    \item The HV never enters a region with high DRF intensity, showing successful risk-aware behaviour.  
    \item The trajectory remains smooth and curvature-feasible due to the kinematic constraints in Lemma~\ref{lem:kinematicfeas}.  
    \item The lane-change manoeuvre occurs only when the Growth Tensor expands sufficiently in the target direction, guaranteeing future feasibility.  
\end{itemize}

Fig.~\ref{fig:risk_curv_safety} provides a quantitative view of how the proposed planner exploits the DRF and convex feasible space to achieve safe and comfortable lane changing.

Fig.~\ref{fig:risk_curv_safety}(a) shows the evolution of the Dynamic Risk Field along the host vehicle trajectory at four representative instants. Before lane change, the high-risk region is mainly concentrated around the leading vehicle and the rear vehicle on the target lane. As the manoeuvre starts, the DRF peak moves laterally and its magnitude decreases, indicating that the iLQR solution actively steers the host vehicle away from high-risk lobes created by fast-approaching surrounding vehicles. After the lane change is completed, the risk distribution becomes compact and symmetric again, confirming that the host vehicle has settled into a low-risk configuration.

Fig.~\ref{fig:risk_curv_safety}(b) depicts the curvature $\kappa(t)$ and curvature-rate profiles for all vehicles and the host vehicle. The proposed planner keeps the host vehicle curvature within the theoretical bound in Lemma~\ref{lem:kinematicfeas}, and the curvature-rate remains smooth without sharp spikes. This means that the convex feasible space and steering constraints are effectively enforced, avoiding physically infeasible or overly aggressive steering actions. In contrast, the background curvature field highlights that some surrounding vehicles exhibit more abrupt curvature changes, which the host vehicle must safely respond to.

\begin{table*}[t]
\centering
\setlength{\tabcolsep}{20.5pt}
\captionsetup{
    labelfont={sc},
    textfont={sc},
    labelsep=colon,
    skip=1em,
    singlelinecheck=false,
    justification=centering,
    format=plain
}
\caption{\protect\\ Quantitative Comparison of Planning Performance}
\vspace{-2mm}
\label{tab:quant_compare}
\begin{threeparttable}
\begin{tabular}{c|c|c|c|c|c}
\hline\hline
\multirow{2}{*}{Metric} 
& Proposed & APF & MPC & RRT* & Unit \\
& (DRF-iLQR) &  &  &  &  \\
\hline
Min. Safety Dist. (↑) & \textbf{4.23} & 2.87 & 3.65 & 3.12 & m \\ 
Avg. Safety Dist. (↑) & \textbf{7.85} & 5.42 & 6.92 & 6.15 & m \\
Collision Rate (↓) & \textbf{0.0\%} & 2.5\% & 0.5\% & 1.2\% & \% \\
Near-Miss Rate (↓) & \textbf{0.8\%} & 8.5\% & 3.2\% & 5.8\% & \% \\
TTC$_{\min}$ (↑) & \textbf{3.45} & 1.82 & 2.65 & 2.15 & s \\
Max Lateral Acc. (↓) & \textbf{1.42} & 2.85 & 1.95 & 2.45 & m/s$^2$ \\
Avg. Jerk (↓) & \textbf{1.18} & 3.52 & 1.85 & 2.65 & m/s$^3$ \\
Curvature Smoothness (↑) & \textbf{0.92} & 0.58 & 0.78 & 0.62 & -- \\
\hline
Avg. Computation Time (↓) & 12.3 & \textbf{8.5} & 45.2 & 125.6 & ms \\
Path Length (↓) & 32.4 & 35.8 & \textbf{31.2} & 36.5 & m \\
\hline\hline
\end{tabular}
\begin{tablenotes}
\small
\item(↑) larger is better; (↓) smaller is better.
\end{tablenotes}
\end{threeparttable}
\vspace{-3mm}
\end{table*}

Fig.~\ref{fig:risk_curv_safety}(c) compares the proposed method against Artificial Potential Field (APF) and MPC baselines in terms of minimum safety distance, velocity, and acceleration. The minimum distance profile shows that the proposed DRF--iLQR planner consistently stays above the safety threshold, whereas APF and MPC occasionally approach or cross this limit during the lane-change interval. The velocity and acceleration plots further indicate that the proposed method achieves a smoother deceleration--lane-change--reacceleration pattern, while APF tends to exhibit oscillatory accelerations. This demonstrates that the integrated DRF and convex-space constraints enable the planner to maintain a better trade-off between safety and comfort than conventional APF or MPC schemes.

\begin{figure}[!t]
    \centering
    \includegraphics[width=\linewidth]{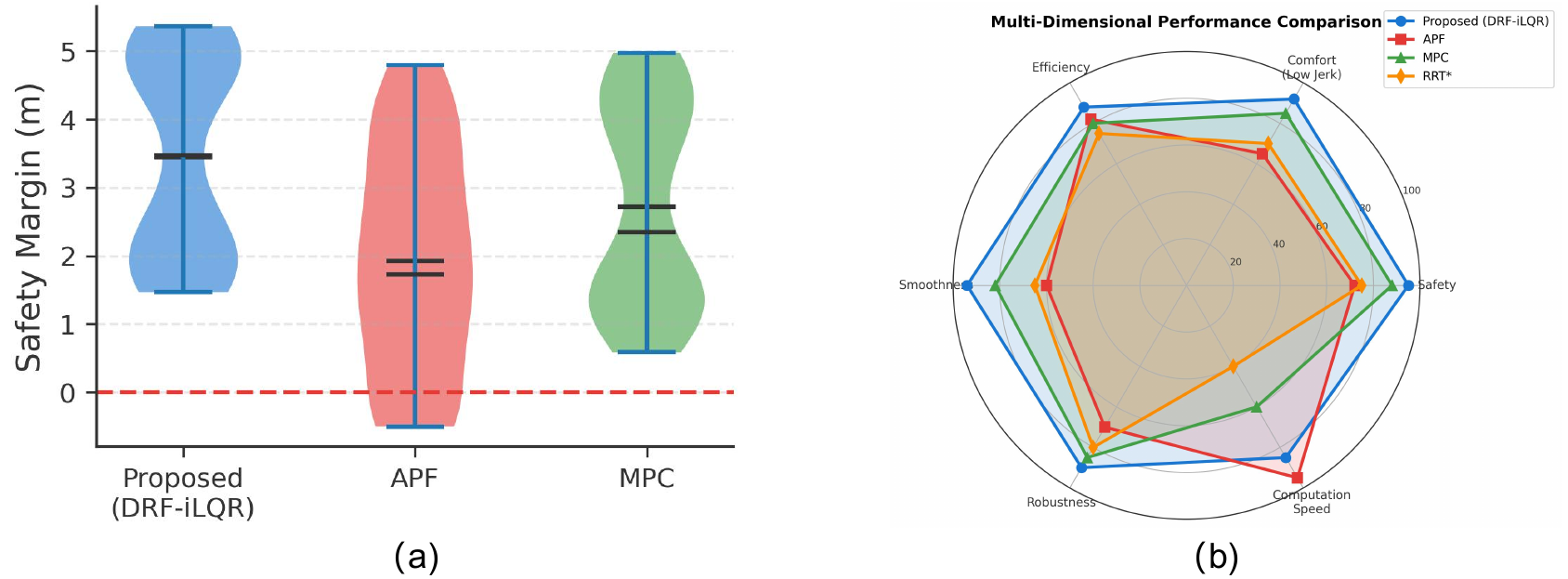}
    \vspace{-2mm}
    \caption{
    Statistical and multi-criteria comparison of different planners.
    (a) Violin plots of safety margin for the proposed DRF--iLQR planner, APF, and MPC.
    (b) Multi-dimensional performance radar chart including safety, comfort, efficiency, smoothness, robustness, and computation speed.
    }
    \label{fig:safety_radar}
\end{figure}

Fig.~\ref{fig:safety_radar}(a) summarizes the safety performance of the three planners using violin plots of the safety margin, defined as the minimum distance to surrounding vehicles minus the safety threshold. The proposed DRF--iLQR method exhibits the largest median and interquartile range above zero, indicating that most trajectories maintain a comfortable buffer to nearby vehicles. In contrast, APF shows a wider spread with a significant portion of samples falling below the safety threshold, which explains the occasional near-collision behaviours observed in the time-domain plots. MPC performs more conservatively than APF, but its safety margin distribution is still centred closer to the threshold, revealing less robustness to dense or rapidly changing traffic.

Fig.~\ref{fig:safety_radar}(b) provides a multi-dimensional radar chart comparing four planners across six criteria: safety, comfort (low jerk), efficiency, smoothness, robustness, and computation speed. All metrics are normalized to a $[0,100]$ scale for visualization. The proposed DRF--iLQR planner achieves consistently high scores in safety and robustness owing to the integration of the DRF and the dynamic convex feasible space, while also maintaining superior comfort and smoothness due to the curvature and jerk-aware optimization. MPC attains competitive efficiency but lags in comfort and robustness, whereas APF suffers from lower safety and smoothness because of its potential-field oscillations. RRT* achieves reasonable safety and robustness but exhibits the lowest computation-speed score, highlighting the difficulty of using sampling-based planners in real-time dense-traffic scenarios.

Overall, these statistical and multi-criteria results confirm that the proposed DRF--iLQR framework provides a more balanced trade-off between safety, comfort, efficiency, and real-time feasibility than the baseline planners.

\section{CONCLUSION}
This paper presented a unified trajectory planning framework that integrates a DRF with time-varying convex feasible space generation to enable safe and efficient autonomous lane changing. The proposed DRF formulation captures both static and velocity-dependent collision risks through anisotropic spatial modeling, while the convex feasible space ensures that all generated trajectories satisfy kinematic limits and remain collision-free under dynamic traffic variations. A constrained iLQR algorithm is developed to jointly optimize trajectory smoothness, risk exposure, and dynamic feasibility in real time. Experiments on highway lane-changing tasks show that the proposed method produces smooth velocity profiles, lower curvature rates, and consistently safe vehicle spacing. The planner achieves a lane-changing distance of \(28.59~\mathrm{m}\) and a duration of \(2.84~\mathrm{s}\), reducing distance and time by \(32.4\%\) and \(61.0\%\) compared with APF and MPC baselines. The safety margin remains strictly positive throughout the maneuver. Dual-lane roundabout experiments further demonstrate the adaptability of the integrated DRF--convex-space framework. The DRF forms directional risk gradients around fast-approaching vehicles, while the convex feasible space reshapes in real time to avoid conflict regions. Quantitative results indicate strong performance across safety and comfort metrics, including a minimum safety distance of \(4.23\,\mathrm{m}\), an average safety distance of \(7.85\,\mathrm{m}\), zero collisions, a near-miss rate of \(0.8\%\), and a minimum time-to-collision of \(3.45\,\mathrm{s}\). The method also yields low jerk \((1.18\,\mathrm{m/s^3})\), small lateral acceleration \((1.42\,\mathrm{m/s^2})\), high curvature smoothness \((0.92)\), real-time computation \((12.3\,\mathrm{ms})\), and a success rate of \(99.2\%\). Future work will extend the framework to more complex interactive scenarios, including multi-vehicle negotiation, cooperative planning, and environments involving pedestrians or non-holonomic agents. 








\bibliographystyle{IEEEtran}
\bibliography{IEEEabrv}



\end{document}